\documentclass[10pt,twocolumn,letterpaper]{article}
\usepackage{float}
\usepackage[caption = false]{subfig}
\usepackage[final]{graphicx}
\usepackage{cvpr}
\usepackage{soul}
\usepackage{times}
\usepackage{epsfig}
\usepackage{graphicx}
\usepackage{amsmath}
\usepackage{enumitem}
\usepackage{amssymb}
\usepackage{float}
\usepackage{tabularx}
\usepackage{caption}
\usepackage{multirow}
\usepackage{graphicx}
\usepackage{xcolor}

\cvprfinalcopy 

\def\cvprPaperID{****} 

\begin{document}

\title{Understanding Knowledge Gaps in Visual Question Answering: Implications for Gap Identification and Testing} 

\author{Goonmeet Bajaj\\
The Ohio State University\\
{\tt\small bajaj.32@osu.edu}
\and Bortik Bandyopadhyay \\ The Ohio State University\\ {\tt\small bandyopadhyay.14@osu.edu}
\and Daniel Schmidt \\ Wright State University\\ {\tt\small schmidt.152@wright.edu}
\and Pranav Maneriker \\ The Ohio State University\\ {\tt\small maneriker.1@osu.edu}
\and Christopher Myers \\ AFRL\\ {\tt\small christopher.myers.29@us.af.mil}
\and Srinivasan Parthasarathy \\ The Ohio State University\\ {\tt\small srini@cse.ohio-state.edu}
}

\maketitle

\begin{abstract}
Visual Question Answering (VQA) systems are tasked with answering natural language questions corresponding to a presented image. Traditional VQA datasets typically contain questions related to the spatial information of objects, object attributes, or general scene questions. Recently, researchers have recognized the need to improve the balance of such datasets to reduce the system's dependency on memorized linguistic features and statistical biases, while aiming for enhanced visual understanding. However, it is unclear whether any latent patterns exist to quantify and explain these failures. As an initial step towards better quantifying our understanding of the performance of VQA models, we use a taxonomy of Knowledge Gaps (KGs) to tag questions with one or more types of KGs. Each Knowledge Gap (KG) describes the reasoning abilities needed to arrive at a resolution. After identifying KGs for each question, we examine the skew in the distribution of questions for each KG. We then introduce a targeted question generation model to reduce this skew, which allows us to generate new types of questions for an image. These new questions can be added to existing VQA datasets to increase the diversity of questions and reduce the skew.
\end{abstract}

\section{Introduction}
When compared to artificially intelligent (AI) systems, human cognition demonstrates a reasonably flexible system when faced with gaps in knowledge while executing a prescribed task (i.e., humans tend not to halt). Humans often demonstrate not only the identification of a gap in knowledge but also the ability to resolve these gaps through various means (question asking, research, \etc).
The ability to identify and resolve knowledge gaps (KGs) provides mechanisms that promote increased flexibility and robustness when faced with imperfect or incomplete information. Informally, a knowledge gap (KG) is an instance of limited or missing information or capabilities, which leads to an agent being inefficient or incapable of completing a given task.

Similar to humans, AI agents do not always have perfect knowledge of a specified task \cite{yang2018visual}. For example, an agent might fail to recognize some objects in an image. Therefore, a framework for KG identification and resolution can facilitate flexibility for an AI agent during both training and execution.
To understand how to build a framework for AI agents to detect, identify, and resolve the different types of KGs that can occur, we use {\it{visual question answering}} (VQA) tasks. VQA sits at an intersection of three components of artificial intelligence: {\it{language}}, {\it{vision}}, and {\it{reasoning}}, thus making it a challenging task. To the best of our knowledge, we are the first to systematically manipulate VQA questions to produce knowledge gaps within AI agents.

\begin{figure}[!t]
\begin{center}
  \includegraphics[width=\linewidth]{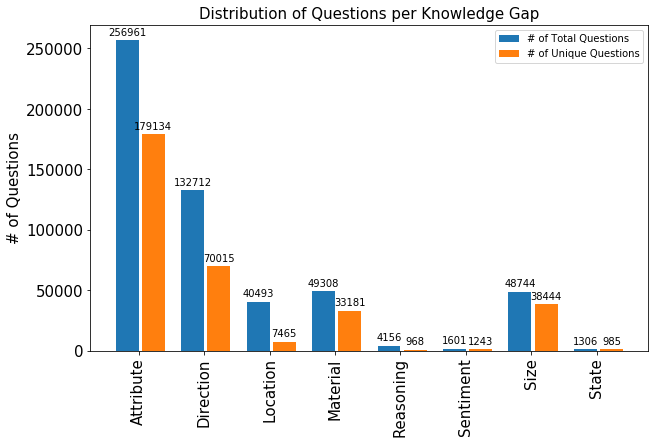}
\end{center}
  \caption{Distribution of Questions per KG}
\label{fig:kg_distr}
\end{figure}

Using a refined version of a KG taxonomy \cite{schmidt2020thesis}, we identify eight different KGs that occur in the GQA dataset \cite{hudson2019gqa}.
Figure \ref{fig:kg_distr} shows the skew we observe in the distribution of the number of questions per KG category.
To alleviate this skew and make questions more evenly distributed across KGs, we apply a neural framework to generate questions for specific KGs.

In Section \ref{rel_works}, we present related works for VQA and natural language question generation. Section \ref{kg_Defin_sec}, contains the definitions of the different types of KGs that we have observed in the VQA setting. Section \ref{kg_iden_sec} contains the details of our KG tagging methodology used for each question. Section \ref{ques_gen_section} defines our neural question generation framework. Section \ref{on_going_and_challenges} describes our on-going work and challenges we face.
\section{Related Work}
\label{rel_works}
In Section \ref{rel_work_vqa}, we explain current efforts in the VQA domain to improve the robustness of VQA models. In Section \ref{quest_gen_rel_work}, we discuss the question generation strategies that lead us to our current approach.


\subsection{Commonsense Reasoning for VQA Models}
\label{rel_work_vqa}

Recent VQA datasets are a result of the realization that VQA models can be highly contingent upon statistical biases and tendencies of the answer distribution and linguistic features \cite{goyal2017making, agrawal2016analyzing, kafle2017visual}. Researchers have pointed out that these biases often result in high accuracies of the VQA models even if image features are ablated. These higher numbers incorrectly lead us to believe that current techniques are making significant progress towards the visual understanding of images. These insights have led to the augmentation of traditional VQA datasets to make VQA models more robust.

Authors of the VQA 2.0 dataset \cite{goyal2017making} augment the VQA \cite{antol2015vqa} dataset with additional images to improve the entropy of answers by creating a uniform distribution of answer probabilities. They \cite{goyal2017making} collect complementary images for every question in their balanced dataset to gather a pair of similar images that result in two different answers to the same question. Singh \etal \cite{singh2019towards} augment the existing Open Images v3 dataset \cite{krasin2017openimages} to increase question and image pairs that require reasoning over the text present in images. They aim to increase the dependency of VQA models on the text present in images by filtering for image and question pairs in which the answer requires the use of OCR text to answer the question. The authors of the GQA dataset \cite{hudson2019gqa} use the Visual Genome Scene Graphs \cite{krishna2017visual} to create a new dataset. The goal of this work is to gain control over the answer distribution and mitigate the heavy dependence on linguistic features and priors in current VQA frameworks. Moreover, this dataset contains rich image annotations using scene graphs and question-answer annotations for common day-to-day images.

The OK-VQA dataset \cite{marino2019ok} contains examples in which the image content is not sufficient to answer the questions. The authors show that state-of-the-art VQA models perform poorly with their new dataset. Shah \etal \cite{shah2019kvqa} release a new dataset, KVQA, that requires compositional reasoning based on vision and commonsense. KVQA contains image and question pairs regarding named entities. The questions in their dataset also require multi-entity, multi-relation, and multi-hop reasoning over large knowledge graphs. Gao \etal \cite{gao2019two} release a new dataset in which they automatically generate compositional questions using both the scene graph of a given image and an external knowledge graph. Their goal was to create a dataset that requires commonsense and visual reasoning. The FVQA \cite{wang2018fvqa} and KB-VQA \cite{wang2015explicit} datasets primarily contain questions that require external information to answer them.  The FVQA dataset contains both image-question-answer triplets and the supporting fact (from an external KB) that is used to answer the question.

These datasets aim to increase the robustness of VQA models and improve their usability. Our work augments the GQA dataset and provides reasoning abilities needed to answer VQA questions. We choose to work with the GQA dataset because contains both image annotations (scene graphs) and question annotations. To the best of our knowledge, we are the first to tag questions with the reasoning skills required to answer them. Our approach provides a new way to analyze the performance of VQA models using different KG categories.

\subsection{Question Generation}
\label{quest_gen_rel_work}
Upon tagging questions with KGs, we observe a skew in the distribution of questions for each KG type. To overcome this skew, we generate new question templates and populate them using image annotations to create new questions for images.
Our question generation procedure is inspired by \cite{serban2016generating, liu2017large}. Serban \etal \cite{serban2016generating} generate question and answer pairs using a neural network architecture to transduce Knowledge Base (KB) facts into natural language questions. The authors use a sequence-to-sequence (seq2seq) framework to generate questions using $(subject,\ predicate,\ object)$ triples from their KB. Similarly, Liu \etal \cite{liu2017large} propose a neural network architecture that combines template-based question generation techniques and seq2seq learning approaches to generate new questions. The authors first create question templates by replacing the subject of the question with a ``SUB" placeholder token. They use these training question templates and KB triples to generate new question templates. Later, they populate the generated question templates with facts from KB triples. We closely follow the approach in \cite{liu2017large} to generate question templates that complement the existing questions in the GQA dataset. However, we expand the question generation framework to use paths in an image's scene graph. We populate these newly generated question templates with information from the scene graph path used to generate the question template. Additionally, since our framework is not restricted to using triples (from a scene graph) to generate new question templates, we can generate questions templates for objects connected via other intermediate objects.


\begin{figure*}[!ht]
\begin{center}
\includegraphics[scale=0.75,width=\linewidth]{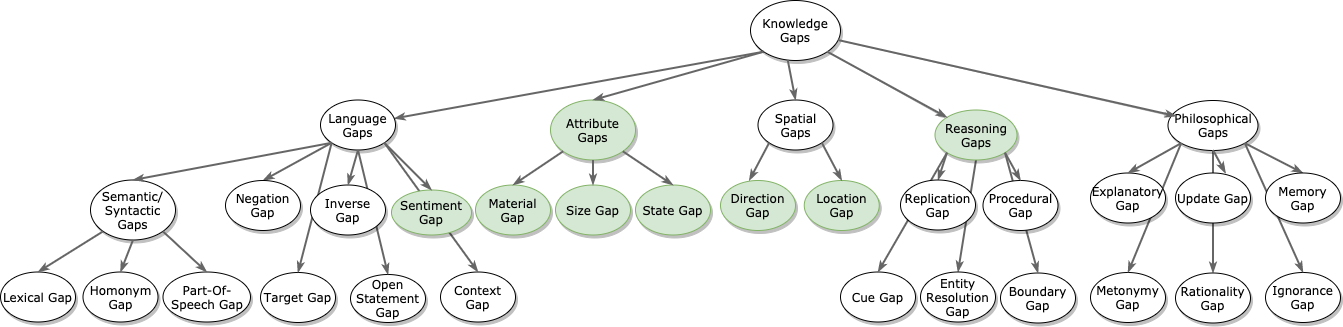}
\end{center}
   \caption{Knowledge Gap Taxonomy}
\label{fig:knowledg_gap_tax}
\end{figure*}

\section{Definition of Knowledge Gaps}
\label{kg_Defin_sec}

In the VQA setting, KGs can occur when an AI model or a human cannot answer a question corresponding to a presented image. This failure can occur for a variety of reasons including, but not limited to, missing information, lack of reasoning skills, and incorrect questions. KGs can occur both at a global or local level. Global gaps are those in which the ground truth VQA response is both unknown to the machine and the oracle. For example, a global KG occurs when the following image and question pair are presented to an agent:
\begin{itemize}[noitemsep]
    \item[] \textbf{Image description}: a man drinking out of a cup with its contents hidden from view
    \item[] \textbf{Question}: {\it{``What is the man drinking?"}}
\end{itemize}
Here, humans might be able to make an educated guess, such as {\it{coffee}} or {\it{tea}}, based on the cup's shape. However, there is no correct answer based on just the image. Local knowledge gaps occur when the correct response exists in some format and can be obtained. For example, humans are faced with this problem when given a novel lexical item. Similarly, for machines, this gap can also occur when VQA models come across words that do not exist in their vocabulary. Using the refined KG taxonomy presented in Figure \ref{fig:knowledg_gap_tax},
we propose to identify the different types of local gaps that can occur when VQA models are presented with a question and image pair. Tagging questions with KGs allows us to understand the difficulty of a dataset in terms of the cognitive skills required to answer questions.

There are five major types of KGs in the taxonomy presented in Figure \ref{fig:knowledg_gap_tax}, namely: {\it{Language}}, {\it{Spatial}}, {\it{Attribute}}, {\it{Reasoning}}, and {\it{Philosophical}} Gaps. \textbf{Language gaps} arise when unknown phrases or vocabulary words are introduced. \textbf{Spatial gaps} occur when there is an error in understanding the physical space of a given setting. \textbf{Attribute gaps} can occur when an object's (or person's) characteristics are not well understood. \textbf{Reasoning gaps} indicate that an agent has difficulty in the cognitive process of understanding information. \textbf{Philosophical gaps} are similar to reasoning gaps but require meta-cognitive processes. We refer the readers to \cite{schmidt2020thesis} for a definition of all KGs presented in the taxonomy.

In this work, we focus on eight KGs (colored in Figure \ref{fig:knowledg_gap_tax}): {\it{Attribute}}, {\it{Direction}}, {\it{Location}}, {\it{Material}}, {\it{Reasoning}}, {\it{Sentiment}}, {\it{Size}}, and {\it{State}} Gaps. Additionally, we can simulate other KGs (e.g., {\it{Explanatory}}, {\it{Context}}, and {\it{Inverse}} Gaps) for the GQA dataset. We leave the simulation of these KGs as future work and describe ongoing efforts in Section \ref{kg_simulation}.

\begin{figure*}[!h]
 \subfloat[\textbf{Q}: Are there any large mouse pads? \newline   \textbf{KG}: Attribute, Size \newline \textbf{Q}: What kind of device is to the left of the cup on the right?  \newline \textbf{KG}: Direction \label{subfig-1:desk}]{%
   \includegraphics[scale=0.75]{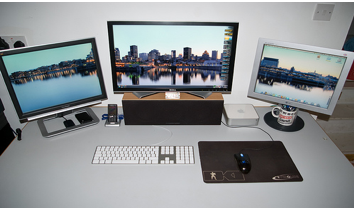}
 }
 \hfill
  \subfloat[\textbf{Q}: Which are less healthy, the brownies or the cherries? \newline  \textbf{KG}: Reasoning \label{subfig-2:dummy}]{%
   \includegraphics[scale=0.75]{2335010.png}
 }
 \hfill
 \subfloat[\textbf{Q}: What is the device that the happy man is holding? \newline  \textbf{KG}: Sentiment \label{subfig-2:dummy}]{%
   \includegraphics[scale=0.75]{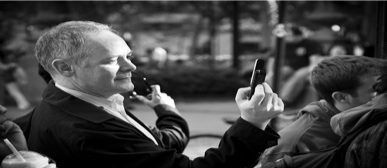}
 }
 \hfill
  \subfloat[\textbf{Q}: Is the large propeller blue and still?  \newline  \textbf{KG}: Attribute, Size, State \label{subfig-2:dummy}]{%
   \includegraphics[scale=0.75]{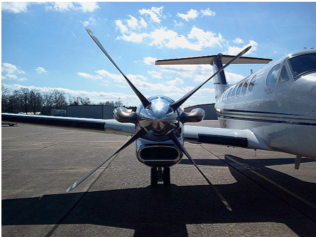}
 }
 \hfill
    \subfloat[\textbf{Q}: Which material makes up the white sink, porcelain or chrome? \newline  \textbf{KG}: Material, Attribute \label{subfig-2:dummy}]{%
   \includegraphics[scale=0.25]{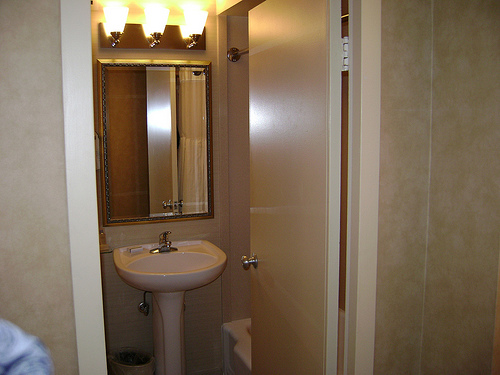}
 }
 \hfill
   \subfloat[\textbf{Q}: Where is the horse that looks white and brown walking? \newline  \textbf{KG}: Attribute, Location \label{subfig-2:dummy}]{%
   \includegraphics[scale=0.75]{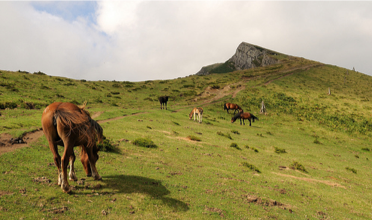}
 }
 \caption{Sampled Images and Questions from the GQA Dataset \cite{hudson2019gqa} with KG Tags}
 \label{fig:Example Images}
\end{figure*}

We now define a subset of the KGs and ground them for the VQA setting. Figure \ref{fig:Example Images} contains sample images and questions from the GQA dataset  \cite{hudson2019gqa}. Under each image in the figure, we present the sampled questions from the dataset and their corresponding KGs tags assigned by our KG tagging methodology (described in Section \ref{kg_iden_sec}).

\smallskip
\noindent {\bf Attribute Gap:}
In the VQA setting, questions inquiring about the color or shape of an object in an image are marked with attribute gap.

\smallskip
\noindent {\bf Context Gap:}
A context gap can occur if the agent has no information about a concept but is aware that such a concept exists. In Section \ref{kg_simulation}, we outline a method to generate these gaps.

\smallskip
\noindent {\bf Direction Gap:}
Direction gaps can arise when agents are not able to understand the differences in spatial relations, such as {\it{left of}}, {\it{right of}}, {\it{near}}, or {\it{beneath}}. In the VQA setting, questions inquiring about relative positions of objects are tagged with a direction gap.

\smallskip
\noindent {\bf Entity Resolution Gap:}
If an agent is unable to identify the correct entity mentioned in the related question or task, then it is encountering an entity resolution gap. These gaps can be simulated for the VQA setting and this procedure is outlined in Section \ref{kg_simulation}.

\smallskip
\noindent {\bf Explanatory Gap:}
Explanatory gaps occur when an agent understands something but is unable to explain {\it{how}} or {\it{why}} it occurs (or is). These gaps can be simulated for the VQA setting. We describe this procedure in Section \ref{kg_simulation}.

\smallskip
\noindent {\bf Inverse Gap:}
Inverse gaps can occur if an agent does not understand the inverse of a relationship or concept. Section \ref{kg_simulation} describes the procedure to simulate these gaps.

\smallskip
\noindent {\bf Location Gap:}
Location gaps can occur when there is a misunderstanding about a specific physical place or setting of a context. Questions in the VQA datasets inquiring about the location of an image scene are marked with this gap. For the GQA dataset, we separate location gaps from direction gaps to create a distinction between questions about an image's scene and questions about the spatial relationships of objects in an image.

\smallskip
\noindent {\bf Material Gap:}
We define a material gap as a subtype of Attribute gap, as this gap focuses on understanding the composition of objects (i.e., wooden, metallic, fabric, and \etc). Questions inquiring about the material or composition of objects in an image are marked with a material gap.

\smallskip
\noindent {\bf Reasoning Gap:}
Questions that require external knowledge about the scene or the objects in an image are marked with a reasoning gap.

\smallskip
\noindent {\bf Sentiment Gap:}
A sentiment gap can occur when an agent is not able to understand the emotion or attitude of another agent. Questions inquiring about the sentiment of an object (typically humans or animals) in an image are marked with a sentiment gap.

\smallskip
\noindent {\bf Size Gap:}
Size gaps, a subtype of Attribute gap, can occur when an agent is trying to understand the physical space an object or a person occupies. Questions inquiring about the size, age, or height of an object in an image are marked with a size gap.

\smallskip
\noindent {\bf State Gap:}
State gaps are a subtype of Attribute gap. These gaps can occur when an agent is trying to understand the specific condition of an object or a person. Questions inquiring about the status of an object in an image are marked with a state gap.

We use these KG definitions to create a rule-based tagging system to mark questions with their respective KGs automatically. This rule-based tagging system is described in the next section.


\section{Knowledge Gap Identification}
\label{kg_iden_sec}

The GQA dataset \cite{hudson2019gqa} consists of 22M questions about various day-to-day images. There are about 113K images in the dataset, 1878 possible answers, and a vocabulary of 3097 words. Many questions in the dataset require multiple reasoning skills, including spatial understanding and multi-step inference. The authors balance the dataset by controlling the answer distribution for various collections of questions, to limit educated guesses using language and world priors. The dataset contains images, scene graphs, questions, object features, and spatial features. For each image, the scene graph contains the image's objects, object attributes, and relations among objects. Figure \ref{fig:Sample Scene Graph} visualizes part of a scene graph. Moreover, questions in the dataset are annotated with a functional program. Functional programs are structured representations that model the semantics of the reasoning steps required to answer the question. The dataset is split into three sets: training, validation, and test. We choose to work with the training dataset and direct the readers to the official GQA website\footnote{\url{https://cs.stanford.edu/people/dorarad/gqa/about.html}} for more information about the original dataset. This section explains the method used to identity the KGs for each question and the challenges we face.


\begin{figure}
\begin{center}
\includegraphics[width=\linewidth]{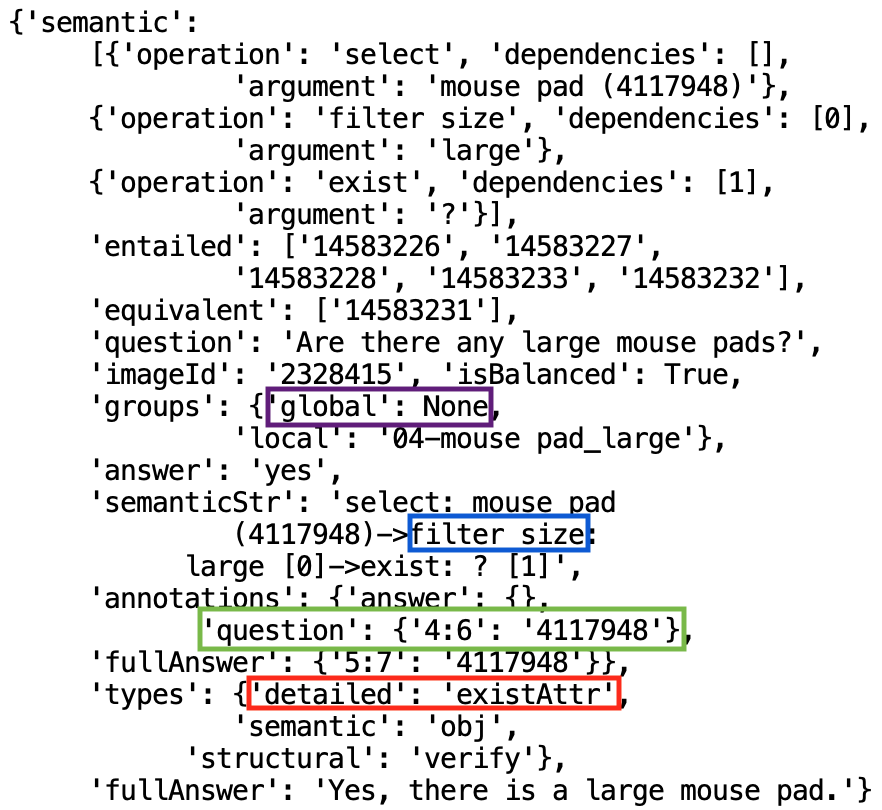}
\end{center}
  \caption{Sample question annotation for Figure \ref{subfig-1:desk}}
\label{fig:question_anno_example}
\end{figure}

\begin{table*}[]
\begin{tabularx}{\textwidth}{|p{1.5cm}|p{6.25cm}|p{4cm}|X|}
 \hline
 \textbf{KG} &
  \textbf{Detailed Types} &
  \textbf{Global Group Label} &
  \textbf{Semantic Filters}\\
  \hline
   Attribute &
  existAttr, existAttrOrC, existAttrNot, verifyAttrK, chooseAttr, verifyAttrs, verifyAttrC, verifyAttr, verifyAttrThis, existAttrC, existAttrOr, categoryAttr, verifyAttrsC, verifyAttrCThis, existAttrNotC, verifyAttrAnd, verifyAttrKC &
  color, shape &
  color, shape \\ \hline
  Direction &
  dir, positionVerify, positionVerifyC, positionChoose,  positionQuery &
  &
  hposition, vposition \\ \hline
Location &
  place, placeVerify, placeVerifyC,  placeChoose, locationVerifyC, locationVerify &
  place, room, nature environment, urban environment, road &
  location, place, room \\ \hline
Material &
  twoSameMaterial, sameMaterialRelate, materialChoose, verifyMaterialAnd, twoSameMaterialC, existMaterialNot, existMaterialNotC, existMaterialC, existMaterial, materialVerify, materialVerifyC, material &
  material, ingredient, texture, textile, liquid, brightness, opaqness\footnote{This is the spelling used in the dataset.}, hardness, pattern &
  liquid, opaqness, material, hardness, pattern, brightness \\ \hline
Reasoning &
  diffAnimals, diffAnimalsC, sameAnimals, sameAnimalsC, comparativeChoose &
   &
   \\ \hline
Sentiment &
   &
  face expression &
  face expression \\ \hline
Size &
  &
  age, height, thickness,  depth, fatness, length, weight, width, size &
  age, fatness, length, thickness, size, weight, depth, width, height \\ \hline
State &
  state &
  state &
  state \\ \hline

\end{tabularx}
\captionsetup{justification=centering}
\caption{KG Tagging}
\label{table:question_anno_to_KG}
\end{table*}

The GQA dataset provides a rich set of annotations for each question. Figure \ref{fig:question_anno_example} contains a sample question annotation (from the dataset) that corresponds to the first question of Figure \ref{subfig-1:desk}.
Table \ref{table:question_anno_to_KG} contains the manual mapping we define between parts of a question annotation and a KG. For each KG in this table, we list the keywords for each of the three annotations we found relevant for this mapping: {\it{detailed type}}, {\it{global group}}, and {\it{semantic filters}}).

\smallskip
\noindent
{\it{\textbf{Detailed type:}}} A question's {\it{detailed type}} is derived from the question's {\it{types}} annotation (green box in Figure \ref{fig:question_anno_example}). The detailed type annotation describes the question's structural (e.g., query (open ended), verify (yes/no)), and semantic type (e.g., attribute for questions about an object's color).

\smallskip
\noindent
{\it{\textbf{Global group:}}} A question's {\it{global group}} is derived from the question’s functional program that can be used to answer the question. The functional program for each question lists a series of series of steps needed to arrive at the answer. The global group is assigned based on the question's answer type (e.g. {\it{color}} for {\it{``What color is the apple?"}}).

\smallskip
\noindent
{\it{\textbf{Semantic filters:}}} The {\it{semantic filters}} for a question are extracted from the question's functional program. The functional program for each question lists a series of series of steps needed to arrive at the answer. We use regular expressions to parse for a list of filter expressions and categorize the list for each KG.

We assign KGs to a question in three steps. First, we use the question's {\it{detailed type}} to assign a KG. If a question's {\it{detailed type}} is in the list of {\it{detailed type}} for a particular KG, we assign that KG to the question. For example, if a question has {\it{placeVerify}} as its detailed type, we tag the question with {\it{Location Gap}}. Similarly, next, we examine a question's {\it{global group}} to assign a new KG. Lastly, we attempt to designate a KG using the {\it{semantic filters}} extracted from a question's functional program. During each step of the pipeline, we try to assign one KG and do not reassign previously specified KGs. This ensures that a question cannot be tagged with the same KG more than once, but can be assigned more than one KG.

For example, based on the annotation provided in Figure \ref{fig:question_anno_example}, we would assign the question the following KGs: {\it{Attribute}} (based on the {\it{detailed type}}) and {\it{Size}} (based on the {\it{semantic filters}}).

Figure \ref{fig:kg_distr} shows the distribution of the number of questions per KG.  The blue columns describe the number of total questions per KG. The orange columns represent the number of unique question strings per KG.  We provide the orange columns because, in the GQA dataset, a question can be asked for multiple images with varying answers (depending on the image content). We can see that only a fraction of the questions are tagged with reasoning, size, and state gaps, whereas, there are many questions in the dataset about object attributes, material compositions, location of image scenes, and spatial relations of objects. This apparent skew inspires us to use a question generation technique to balance dataset in terms of KGs.

Additionally, we keep track of how each KG tag was assigned (i.e., {\it{detailed type}}, {\it{global group}}, or {\it{semantic filters}}). \autoref{table:KG_Source_Distribution} describes the distribution of different sources of KGs. For example, we see that 114,083 questions are tagged with an attribute gap using their {\it{detailed type}} annotation. Similarly, we can see that no sentiment gaps are assigned using the {\it{detailed type}} annotations. This is consistent with \autoref{table:question_anno_to_KG} as there are no {\it{detailed type}} keywords present in the question annotations to identify sentiment gaps. Instead, we tag 50 questions using the {\it{global group}} label and 1,551 questions using the {\it{semantic filters}} annotations with a sentiment gap.

\begin{table}[]
\begin{center}{%
\scalebox{0.75}{
\resizebox{\linewidth}{!}{%
\begin{tabular}{|l|l|l|l|}
\hline
\textbf{KG} & \textbf{Detailed} & \textbf{Group} & \textbf{Semantic} \\ \hline
Attribute   & 114083            & 55718          & 87160             \\ \hline
Direction   & 98809             & 0              & 33903             \\ \hline
Location    & 22920             & 14712          & 2861              \\ \hline
Material    & 17005             & 8385           & 23918             \\ \hline
Reasoning   & 4156              & 0              & 0                 \\ \hline
Sentiment   & 0                 & 50             & 1551              \\ \hline
Size        & 0                 & 12654          & 36090             \\ \hline
State       & 129               & 120            & 1057              \\ \hline
\end{tabular}%
}}}
\end{center}
\caption{Number of Questions per Source of KG}
\label{table:KG_Source_Distribution}
\end{table}

\section{Question Generation}
\label{ques_gen_section}

We aim to generate new questions for images that lack certain KGs to remove the skew in the distribution of questions for each KG. Note, it may not be possible to generate questions to address all types of KGs for an image. For example, if an image does not contain a human, it is not feasible to generate a question for the sentiment gap. We do not currently handle these cases and leave them as future work. By increasing the number of questions for certain KGs, we can make the dataset more balanced or reduce the skew. For example, sentiment questions require commonsense reasoning. Therefore, if VQA models perform poorly on these types of questions, it may be because the dataset does not contain enough training examples. We do not currently focus on generating the correct answers for the newly generated question and leave this as future work. For completeness, we generate questions for all eight KGs in green in Figure \ref{fig:knowledg_gap_tax}. However, for soundness, we only need to generate questions for the following KGs with a low volume of questions to reduce the skew: Location, Reasoning, Sentiment, and State Gap. Therefore, we bold these KGs in the upcoming Tables  \ref{table:sample_templates},  \ref{table:distribut_training_triple}, \ref{table:distribut_training_path}, \ref{table:case_study_sample_ques_kg}, \ref{table:models_res}.

\begin{figure}[]
\begin{center}
  \includegraphics[width=\linewidth, scale=0.8]{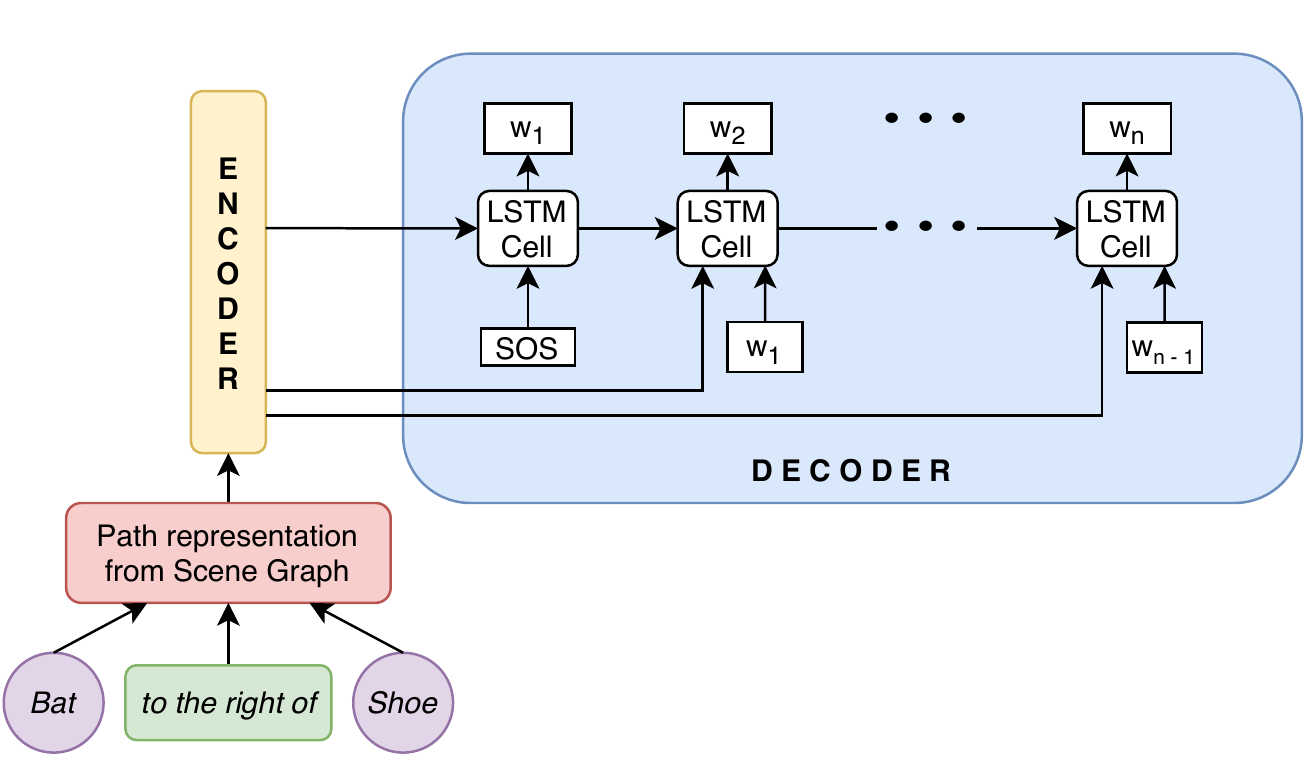}
\end{center}
  \caption{Template Based Seq2Seq Model}
\label{fig:Model}
\end{figure}

Our neural question generation strategy is similar to that of \cite{liu2017large}. We use a neural network architecture that combines template-based question generation methods and seq2seq learning to generate new question templates. We use the IBM Pytorch Seq2Seq framework{\footnote{\url{https://ibm.github.io/pytorch-seq2seq/public/index.html}}}. Figure \ref{fig:Model} depicts the model we use to generate new question templates.

\begin{figure*}[]
\begin{center}
  \includegraphics[width=\linewidth, trim={2cm 2cm 2.2cm 1.5cm}]{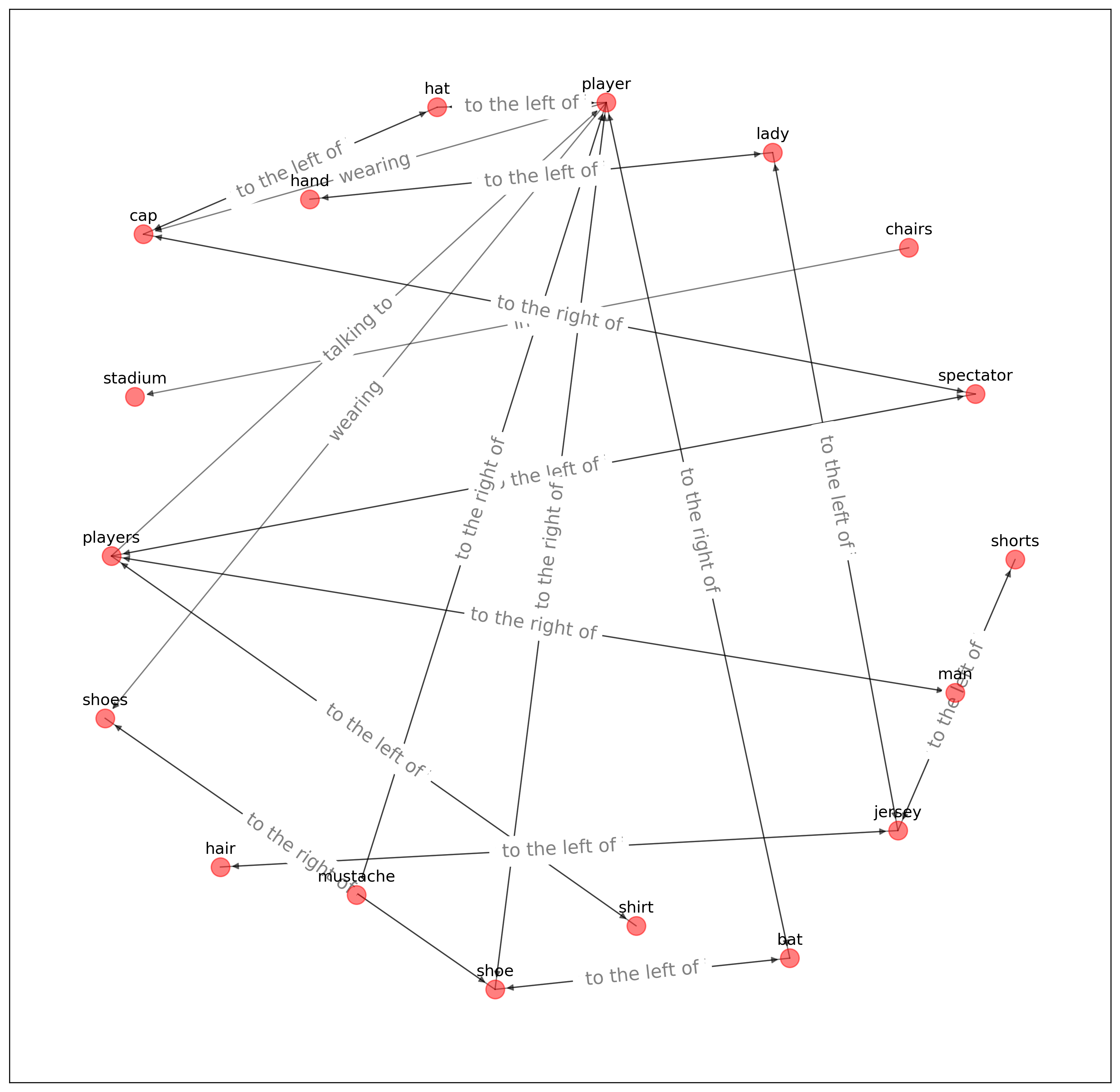}
\end{center}
  \caption{Partial-Sample Scene Graph for Figure \ref{fig:Case_study_image} (object attributes are not visualized)}
\label{fig:Sample Scene Graph}
\end{figure*}

The GQA scene graphs contain objects, their relationships, and their attributes. Figure \ref{fig:Sample Scene Graph} presents a visualized scene graph for Figure \ref{fig:Case_study_image}. In the visualized scene graph, the nodes represent objects in images, and edges represent the relationship among objects.
We aim to generate question templates by extracting paths in a scene graph of length $L$. We have noticed that not all objects in a scene graph are connected (e.g., some nodes do not have an incoming or an outgoing edge). Currently, we do not use these isolated nodes in the scene graph. However, these nodes can be used to generate context gaps (see Section \ref{kg_simulation}).

We can extract a simple path of length $L$ and use it to generate a question template that is concerned with the objects, their attributes, and the relationships along the path sequence. The question template generation method can be modeled in a probabilistic framework \cite{serban2016generating, liu2017large}.

\[P(Q|P) = {\prod_{i=1}^N}
P(w_i | w_{<i}, P) \tag{1} \]
$Q = (w_1, w_2, ... , w_n)$ represents a generated question template that consists of tokens $w_1, w_2, ... , w_n$. In most question templates, the last generated token $w_n$ should be `?'.

\noindent
\textbf{Training path extraction:} $P$ represents a path sequence of length $L$ that is fed into the seq2seq model's encoder. The input path is either a {\it{triple}} ($L$ = 1) or a {\it{simple path}} of length $L$ that is extracted from an image's scene graph based on the corresponding question. Triples are in the form of $P = (g(o_1), r_1, g(o_2))$. The simple paths are represented as $P = (g(o_1), r_1, IO, r_2 , IO, ... , r_n, g(o_2))$. The $o_1$ and $o_2$ are objects mentioned in the training questions. The function $g(\cdot)$ describes objects as a concatenation of their attributes and their names in English. The $r_n$ along a path are the relations among objects. We chose to replace the intermediate objects along a path with a place holder ``IO" (short for ``INTERMEDIATE\_OBJECT"), because we are only interested in objects, $o_1$ and $o_2$, that are actually present in the training question. Additionally, using paths of $L > 1$ can allow us to generate questions about objects that are connected through intermediate nodes. For example, the scene graph for the image in Figure \ref{fig:Case_study_image} does not have a direct edge between the annotated objects: ``player" and ``man". However, if we search for a path between the objects ``player" and ``man" in the scene graph
(Figure \ref{fig:Sample Scene Graph}), we are able to find a relational path between them.

\noindent
\textbf{Training template generation:} To create training question templates, we replace the objects and attributes mentioned in the training question with ``OBJ" and ``ATTRIBUTE" placeholders. The question annotations in the GQA dataset contain indices to mark which words of the question are objects (see green box in Figure \ref{fig:question_anno_example}). We use these indices to find the object words in the question to replace with the ``OBJ" placeholders. To replace the attributes in the question text, we find the attributes of the objects in the scene graph and replace them with the ``ATTRIBUTE" placeholder in the question text. \autoref{table:sample_templates} presents sample questions and their templates that are used during training. Additionally, the case study in Section \ref{case_study} presents sample inputs to the model and the output and populated templates.

\begin{table*}[]
\begin{tabularx}{\textwidth}{|p{1.5cm}|X|X|}
\hline
\textbf{KG} & \textbf{Original Question}                                  & \textbf{Question Template}                                 \\ \hline
Attribute   & What is the green object on the table made of?              & What is the ATTRIBUTE object on the OBJ made of?           \\ \hline
Direction   & Where is the bird on the branch looking at?                 & Where is the OBJ on the OBJ looking at?                    \\ \hline
\textbf{Location}    & Where are the people to the left of the umbrella sitting?   & Where are the OBJ to the left of the OBJ sitting?          \\ \hline
Material    & Was plastic used to make the cookie on the counter?         & Was ATTRIBUTE used to make the OBJ on the OBJ?             \\ \hline
\textbf{Reasoning}   & Which are healthier, the pizza or the peppers of the pizza? & Which are healthier, the OBJ or the OBJ of the OBJ?   \\ \hline

\textbf{Sentiment} & Is the man that is to the left of the other man both old and happy?        & Is the OBJ that is to the left of the other OBJ both ATTRIBUTE and ATTRIBUTE?  \\ \hline
Size        & Is the truck to the left of the mirror white and small?     & Is the OBJ to the left of the OBJ ATTRIBUTE and ATTRIBUTE? \\ \hline
\textbf{State}     & Is the sheep that is to the right of the other sheep still or is it rough? & Is the OBJ that is to the right of the other OBJ ATTRIBUTE or is it ATTRIBUTE? \\ \hline

\end{tabularx}
\captionsetup{justification=centering}
\caption{Sample Question and Extracted Question Templates for each KG}
\label{table:sample_templates}
\end{table*}

Our template-based seq2seq model can be viewed as a translator that converts structured data (paths along a scene graph) into question templates. The model can be divided into a path encoder and a question template decoder. These generated templates can be populated in a downstream task with data from the path to generate a complete question.

\noindent
\textbf{Encoder:} The (Bi)LSTM encoder encodes a path sequence, $P$, from a scene graph (of an image) into an embedding.

\noindent
\textbf{Decoder:} We use the built-in attention mechanism with an LSTM decoder to make use of the alignment information between scene graph paths and question templates. During decoding, we use teacher forcing (ratio = 0.25) to allow the decoder to learn how to generate question templates. We use the {\it{TopKDecoder}}, which performs a beam search of length $K = 10$. Of the top decoded sequences, we select the template with the highest probability and with the same number of ``ATTRIBUTE" placeholders in the generated template as the original training template. If none of the top $K$ results meet this criterion, we use the template with the highest probability as our output.

\subsection{Experimental Setup}
To generate new question templates, we train separate seq2seq question generation models for each KG type. These separate models allow us to control the types of templates we generate, as we try to remove the skew in distribution presented in \autoref{fig:kg_distr}. For each KG type, we train two seq2seq models: 1) using the triple representations ($L = 1$), 2) using the path representations ($L \leq 5$). We refer to the model that uses triples as the {\it{triple-based model}}, and the model uses paths of length $L$ as the {\it{path-based model}}.

\autoref{table:distribut_training_triple} describes the distribution of the triples dataset used for our first model that uses triples as input to the encoder. \autoref{table:distribut_training_path} describes the distribution of the paths dataset used for second model that uses paths as input to the decoder. We use 80\% of our data for training, 10\% for validation, and 10\% for testing.
These tables also reiterate the fact that questions tagged with certain KGs (e.g., state, reasoning, sentiment) are not well represented in the dataset because they do not have a large number of unique training templates.

\begin{table}[]
\resizebox{\linewidth}{!}{%
\begin{tabular}{|l|l|p{ 1.5 cm}|p{ 2 cm}|p{ 2 cm}|}
\hline
\textbf{KG} & \textbf{Split} & \textbf{\# of Training Examples} & \textbf{\# of Unique Training Templates} & \textbf{\# of Unique Training Triples} \\ \hline
\multirow{3}{*}{Attribute} & train & 94936 & 10921 & 46520 \\ \cline{2-5}
                          & val   & 12727 & 1988  & 9696  \\ \cline{2-5}
                          & test  & 12795 & 2003  & 9701  \\ \hline
\multirow{3}{*}{Direction} & train & 34353 & 4261  & 17151 \\ \cline{2-5}
                          & val   & 4370  & 1096  & 3452  \\ \cline{2-5}
                          & test  & 4367  & 1089  & 3498  \\ \hline
\multirow{3}{*}{\textbf{Location}}  &  train & 1867  & 471   & 1486  \\ \cline{2-5}
                           & val   & 236   & 99    & 218   \\ \cline{2-5}
                           & test  & 236   & 96    & 225   \\ \hline
\multirow{3}{*}{Material}  & train & 17328 & 3886  & 11270 \\ \cline{2-5}
                          & val   & 2196  & 549   & 1873  \\ \cline{2-5}
                          & test  & 2191  & 563   & 1951  \\ \hline
\multirow{3}{*}{\textbf{Reasoning}} & train & 659   & 36    & 392   \\ \cline{2-5}
                           & val   & 130   & 21    & 92    \\ \cline{2-5}
                           & test  & 129   & 19    & 101   \\ \hline
\multirow{3}{*}{\textbf{Sentiment}} & train & 1066  & 309   & 822   \\ \cline{2-5}
                           & val   & 135   & 53    & 122   \\ \cline{2-5}
                           & test  & 136   & 59    & 128   \\ \hline
\multirow{3}{*}{Size}      & train & 21925 & 3336  & 14165 \\ \cline{2-5}
                           & val   & 2822  & 737   & 2458  \\ \cline{2-5}
                           & test  & 2828  & 762   & 2445  \\ \hline
\multirow{3}{*}{\textbf{State}}     & train & 346   & 101   & 287   \\ \cline{2-5}
                          & val   & 44    & 23    & 37    \\ \cline{2-5}
                          & test  & 45    & 20    & 44    \\ \hline
\end{tabular}%
}
\caption{Triples Dataset Distribution}
\label{table:distribut_training_triple}
\end{table}

\begin{table}[]
\resizebox{\linewidth}{!}{%
\begin{tabular}{|l|l|p{ 1.5 cm}|p{ 2 cm}|p{ 2 cm}|}
\hline
\textbf{KG} & \textbf{Split} & \textbf{\# of Training Examples} & \textbf{\# of Unique Training Templates} & \textbf{\# of Unique Training Paths} \\ \hline
\multirow{3}{*}{Attribute} & train & 331827 & 11540 & 296698 \\ \cline{2-5}
                          & val   & 43972  & 2044  & 42048  \\ \cline{2-5}
                          & test  & 44759  & 2024  & 42869  \\ \hline
\multirow{3}{*}{Direction} & train & 111937 & 4349  & 100917 \\ \cline{2-5}
                          & val   & 14533  & 1111  & 13720  \\ \cline{2-5}
                          & test  & 14468  & 1102  & 13721  \\ \hline
\multirow{3}{*}{\textbf{Location}}  & train & 5137   & 473   & 4655   \\ \cline{2-5}
                           & val   & 654    & 99    & 596    \\ \cline{2-5}
                           & test  & 667    & 96    & 613    \\ \hline
\multirow{3}{*}{Material}  & train & 52199  & 4029  & 47126  \\ \cline{2-5}
                          & val   & 6743   & 580   & 6304   \\ \cline{2-5}
                          & test  & 6639   & 620   & 6219   \\ \hline
\multirow{3}{*}{\textbf{Reasoning}} & train & 7116   & 39    & 6388   \\ \cline{2-5}
                           & val   & 1016   & 25    & 964    \\ \cline{2-5}
                           & test  & 932    & 27    & 895    \\ \hline
\multirow{3}{*}{\textbf{Sentiment}} & train & 2984   & 311   & 2667   \\ \cline{2-5}
                           & val   & 403    & 53    & 385    \\ \cline{2-5}
                           & test  & 386    & 57    & 361    \\ \hline
\multirow{3}{*}{Size}      & train & 58969  & 3332  & 51647  \\ \cline{2-5}
                           & val   & 7982   & 834   & 7342   \\ \cline{2-5}
                           & test  & 8180   & 848   & 7566   \\ \hline
\multirow{3}{*}{\textbf{State}}     & train & 587    & 106   & 527    \\ \cline{2-5}
                          & val   & 65     & 21    & 56     \\ \cline{2-5}
                          & test  & 67     & 22    & 53     \\ \hline
\end{tabular}%
}
\caption{Paths Dataset Distribution}
\label{table:distribut_training_path}
\end{table}
We perform a grid search for each KG model and select the best model based on our training loss and the validation loss. We vary the following parameters to tune our question generation models:
    \begin{itemize}[noitemsep]
        \item teacher forcing ratio: 0.2, 0.3, 0.4, 0.5
        \item hidden units: 8, 16, 32, 64
        \item learning rate: 0.0001, 0.001, 0.01
        \item direction of the encoder: unidirectional and bidirectional
        \item size of decoder beam search: 5, 8, 10
    \end{itemize}

\subsection{Experimental Results}
We use BLEU\footnote{\url{https://www.nltk.org/_modules/nltk/translate/bleu_score.html}} and Meteor\footnote{\url{https://www.nltk.org/_modules/nltk/translate/meteor_score.html}} scores to evaluate our question generation model. These metrics are commonly used for Natural Language Generation (NLG) task. BLEU scores are computed based on an average of unigram, bigram, trigram, and 4-gram precision. BLEU scores are relatively easier and faster to calculate when compared to human evaluation. Meteor scores are another metric used for evaluating machine translation models. These scores are claimed to have a better correlation with human judgment when compared to BLEU scores. Meteor scores also take into account word order while evaluating generated text.

\autoref{table:models_res} presents the results of our models. Our reasoning, location, and material gaps question generation models achieve the highest Meteor scores for our triple-based input. The path-based input models achieve a higher Meteor score for Attribute, Direction, and Size gaps. Similarly, our BLEU scores are higher for Attribute, Direction, and Size gap path-based question generation models as well (equal BLEU scores for both State gap question generation models). The State gap models do not perform well. We suspect this is because of the low number of training examples (see Table \ref{table:distribut_training_triple} and Table \ref{table:distribut_training_path}).
\begin{table}[]
\scalebox{1.0}{
\resizebox{\linewidth}{!}{%
\begin{tabular}{|p{1.5cm}|p{1cm}|p{1cm}|p{2cm}|p{2cm}|}
\hline
\textbf{KG} & \textbf{BLEU Score} & \textbf{Meteor Score} & \textbf{\# of Novel Templates} & \textbf{\# of Existing Templates} \\ \hline
\multicolumn{5}{|c|}{\textbf{Triple-based Question Generation Model Results}}                                                                      \\ \hline
Attribute & 0.18  & 0.52 & 64 & 208  \\ \hline
Direction & 0.38  & 0.57 & 31 & 78   \\ \hline
\textbf{Location}    & 0.32 & 0.64 & 35 & 14 \\ \hline
Material & 0.30  & 0.59 & 71 & 39  \\ \hline
\textbf{Reasoning}   & 0.75 & 0.87 & 7 & 5 \\ \hline
\textbf{Sentiment}   & 0.21 & 0.53 & 28 & 6 \\ \hline
Size        & 0.16 & 0.50 & 18 & 78 \\ \hline
\textbf{State} &  0.0 & 0.32 & 16 & 0  \\ \hline
\multicolumn{5}{|c|}{\textbf{Path-based Question Generation Model Results}}                                                                        \\ \hline
Attribute & 0.20  & 0.54 & 70 & 254 \\ \hline
Direction & 0.40 & 0.60 & 16 & 5  \\ \hline
\textbf{Location}    & 0.26 & 0.59 & 38 & 19 \\ \hline
Material & 0.26  & 0.56  & 430 & 455 \\ \hline
\textbf{Reasoning}   & 0.69 & 0.83 & 6 & 9 \\ \hline
\textbf{Sentiment}   & 0.17 & 0.50 & 30 & 8 \\ \hline
Size        & 0.16 & 0.53 & 16 & 5 \\ \hline
\textbf{State} & 0  & 0.27  & 11 & 0   \\ \hline
\end{tabular}%
}}
\caption{Results for Each KG Question Generation Model}
\label{table:models_res}
\end{table}

Our path-based model improves the Meteor score and BLEU score by 3\% and 2\% (respectively) for direction gap with respect to our triple-based model. It does not improve performance for other KGs. This result is plausible because questions regarding spatial relations can benefit from more information about how objects are related to each other in an image (via a scene graph). Similarly, the questions concerned with an object's attribute are more dependent on the triples related to that object.

\subsection{Case Study}
\label{case_study}
\begin{figure}[]
\begin{center}
  \includegraphics[scale=0.5]{2377804.jpg}
\end{center}
  \caption{Image Used for Case Study}
\label{fig:Case_study_image}
\end{figure}

In this section, we use a sample image to motivate our work. The image in Figure \ref{fig:Case_study_image} is paired with the questions in \autoref{table:case_study_sample_ques_kg} in the GQA dataset. Additionally, the {\it{KG(s)}} column in \autoref{table:case_study_sample_ques_kg} presents the KGs assigned for each question by our KG tagging method. From \autoref{table:case_study_sample_ques_kg}, we can see that the questions related to the sampled image are primarily about object attributes. Additionally, there is one question about the location of the image scene and two questions about the spatial relations of the objects present in the image. Our KG tagging system identifies the following set of KGs for the questions: attribute, direction, and location gaps. However, there many more KGs that can be generated. The lack of diversity among the types of KGs for Figure \ref{fig:Case_study_image} is consistent with the skew we observe in Figure \ref{fig:kg_distr}. Therefore, we use our question generation models trained for sentiment gap, reasoning gap, material gap, and size gap to create different types of question templates that can be populated to create new questions for the sampled image.

\begin{table}[]
\resizebox{\linewidth}{!}{%
\begin{tabularx}{\linewidth}{|p{6.0cm}|X|}
\hline
\textbf{Question}                     & \textbf{KG(s)} \\ \hline
Where in this photo are the green chairs, in the top or in the bottom? & Direction, Attribute \\ \hline
What place is this?                   & Location       \\ \hline
Is the bat of the players brown?      & Attribute      \\ \hline
Where are the chairs?                 & Location       \\ \hline
Which color is the lady's hair?       & Attribute      \\ \hline
What kind of furniture is green?      & Attribute      \\ \hline
Which type of clothing is not orange? & Attribute      \\ \hline
What kind of clothing is not orange?  & Attribute      \\ \hline
Which color does the shirt have?      & Attribute      \\ \hline
Are there any black hats or gloves?   & Attribute      \\ \hline
Is the man on the right or on the left side of the image?              & Direction            \\ \hline
Which kind of clothing is orange?     & Attribute      \\ \hline
What is the color of the shoes the player is wearing?                  & Attribute            \\ \hline
What color are the shorts?            & Attribute      \\ \hline
What color is the hat?                & Attribute      \\ \hline
\end{tabularx}
}
\caption{Questions in the Dataset for the Case Study Image}
\label{table:case_study_sample_ques_kg}
\end{table}

Using our KG specific trained question generation models, we can generate new question templates that can be populated to increase the variety of questions. We can generate questions for reasoning, material, size, and sentiment gaps. These gaps do not exist in \autoref{table:case_study_sample_ques_kg}. We now share sample outputs from our triple-based question generation models. We share the scene graph triple that is used as input, the question template generated by the model, and the populated template (based on triple data).

\noindent
New template using our triple-based material gap model:
\begin{itemize}[nosep]
        \item[] INPUT: {\it{cap to the left of pants}}
        \item[] OUTPUT TEMPLATE: ``What is the OBJ near the OBJ made of ?"
        \item[] POPULATED TEMPLATE: ``What is the cap near the pants made of?"
\end{itemize}

\noindent
New template using our triple-based reasoning gap model:
\begin{itemize}[nosep]
        \item[] INPUT: {\it{players to the right of man}}
        \item[] OUTPUT TEMPLATE: ``Which is younger, the OBJ or the OBJ ?"
        \item[] POPULATED TEMPLATE: ``Which is younger, the players or the man?"
    \end{itemize}
\noindent
New template using our triple-based sentiment gap model:
\begin{itemize}[nosep]
        \item[] INPUT: {\it{spectator to the right of cap}}
        \item[] OUTPUT TEMPLATE: ``Is the ATTRIBUTE OBJ to the right of the OBJ ?"
        \item[] POPULATED TEMPLATE: ``Is the happy spectator to the right of the cap?"
    \end{itemize}
New template using our triple-based size gap model:
\begin{itemize}[nosep]
        \item[] INPUT: {\it{bat to the right of shoe}}
        \item[] OUTPUT TEMPLATE: ``How big is the OBJ near the OBJ ?"
        \item[] POPULATED TEMPLATE: ``How big is the bat near the shoe?"
    \end{itemize}

These new questions can be used along with the questions that exist in the GQA dataset. These new questions are not grammatically perfect. In our future work, we aim to borrow ideas from the NLG community to increase the quality of our generated questions.

\section{Ongoing Work and Challenges}
\label{on_going_and_challenges}
In this section, we share our ongoing efforts and initial ideas to generate new KGs for the GQA dataset. Moreover, we also present some of the challenges we face.

\subsection{Additional KG Generation}
\label{kg_simulation}
The method described in Section \ref{kg_iden_sec} is designed to capture eight KGs (i.e., Direction, Location, Material, State, Sentiment, Reasoning, Attribute). As previously mentioned, the following KGs can be generated for the GQA dataset: Context Gap, Entity Resolution Gap, Explanatory Gap, and Inverse Gap. A KG can be generated by asking a question that requires specific cognitive skills to answer it. We now describe our ongoing work to expand the types of KGs that can be found in the VQA setting.

\noindent {\bf Context Gap:} Context gaps can be generated using the scene graphs provided with images. We can identity isolated nodes in scene graphs and generate questions regarding those nodes. This is consistent with the definition of context gaps since the agent knows there is an isolated object but is missing the information to connect the object with the rest of the image content.

\smallskip
\noindent {\bf Entity Resolution Gap:} Questions about an object might be ambiguous if there is more than one object of the same type. For example, the following question, {\it{On which side of the picture is the closed shelf?}} cannot be answered correctly because there can be multiple closed shelves in the image that is associated with the question. These gaps can be identified by counting the number of times the object in the question occurs in the scene graph. This technique is highly dependent on the quality of scene graphs and image annotations.

\noindent {\bf Explanatory Gap:}
Explanatory gap questions can be generated for the VQA dataset. Authors of \cite{gao2019two} aim to generate questions that require external knowledge. They automatically generate compositional questions using an image's scene graph and an external knowledge graph. Similarly, for each object in an image, ConceptNet \cite{speer2017conceptnet} can be queried to search for the {\it{UsedFor}} edge. If such an edge exists, the following template can be used to generate a new question: {\it{What is the OBJ used for?}}. The answers to these questions can also be collected by noting the nodes connecting the image objects with the {\it{UsedFor}} edge. This technique can result in some incorrect questions such as:
\begin{itemize}[noitemsep,topsep=0pt]
    \item What is the wine used for?
    \item What is the dog used for?
    \item What is the ocean used for?
\end{itemize}

\noindent {\bf Inverse Gap:} Assigning an inverse gap tag requires identifying (or creating) an opposite question, $Q'$, that compliments the original question, $Q$. One technique to identify, an $Q'$ is to tokenize the original question $Q$ and use WordNet \cite{fellbaum2012wordnet, miller1998wordnet} to search for antonyms of the words in $Q$. To reduce noise, we can use NLTK \cite{loper2002nltk} to identify the search space of the part-of-speech (POS) tag for each token. Then we can only search for antonyms of verbs, adjectives, determiners, and existentials of each token in Q. The WordNet query (through NLTK) can be refined using the token and its coarse-grain POS tag. For all synsets returned for a token, their lemmas' can be used to find all antonyms. For each antonym, we can create a $Q'$ if the antonym is not already present in $Q$. We observe that this last check is necessary because there are ``comparative" questions in the dataset, e.g., {\it{``Is the table small or large?"}}. For these types of questions, it would not make sense to generate an inverse question $Q'$ that asks: {\it{``Is the table large or large?"}}. We can curate all possible inverse questions using this method and further prune the list by selecting all generated questions that exist in the dataset. This rule-based pruning technique removes questions like: {\it{``Is the ball to the wrong of the bat?"}} or {\it{``Is the table big or large?"}} The pruning technique is crafted by observing trends in the generated inverse questions. Finally, the remaining generated inverse questions can then be assigned paired with the respective original question's images and be tagged with an inverse gap.

\subsection{Question Generation}
Our question generation framework can generate question templates given a triple or a path from a scene graph. To make use of these templates, we can use an image's annotation to replace the place holder text with the image's data. We plan to improve the quality of our generated templates by using ideas from the NLG community. Additionally, we plan to try different neural models to generate questions and answers that can be used during training. We also aim to which types of KGs are suitable for an image before we generate questions. For example, this step can help prevent the generation of sentiment questions for images that do not contain humans.

\subsection{Challenges}

The method described for automatically assigning KGs to questions is not free of errors as it can both fail to assign relevant KGs or assign irrelevant KGs. Our current approach is highly dependent on the quality of the question annotations and the handcrafted question-annotations-to-KG mapping. Upon investigation, we have identified several sources of errors, including but not limited to garden path questions, multiple-meaning words (MMW), and dataset annotation errors that can cause problems in the KG tagging process.

\smallskip
\noindent {\bf Garden Path Questions:} Questions in the dataset can be ambiguous due to the lack of punctuation. Therefore, there can be multiple correct answers to a question based on how the agent reads the question, e.g., {\it{``Are both the sofa and the bookcase to the left of the knife made of wood?"}}.

\smallskip
\noindent {\bf Dataset Errors:} Many of the 22 million questions in the dataset are generated through templates. Therefore, errors can occur in question annotations. For example, for the following questions, the {\it{global group}} label in the dataset is {\it{color}}:
\begin{itemize}[noitemsep]
    \item {\it{Which is healthier, the orange or the muffins?}}
    \item {\it{Which is healthier, the candies or the orange?}}
\end{itemize}
This annotation is incorrect because the question is not about the color of an object. Instead, it is asking about the object, {\it{orange}}. Additionally, for some questions, there are parts of annotations that are missing, e.g., {\it{``Is the large pot to the right or to the left of the jar in the middle of the picture?"}}. There is no mention of {\it{size}} or {\it{large}} in the question annotation.

\smallskip
\noindent {\bf Multiple Meaning Word (MMW):} The following questions are tagged with attribute gaps by our system:
\begin{itemize}[noitemsep]
    \item {\it{Which is healthier, the orange or the muffins?}}
    \item {\it{Which is healthier, the candies or the orange?}}
\end{itemize}
Here, the word {\it{orange}} is a noun rather than an adjective. Our method fails to understand that the word {\it{orange}} is not an attribute. Note, this error is also in part because of the dataset annotation errors, as mentioned earlier.

\section{Conclusion}
In this work, we take initial steps to understand the different types of reasoning skills needed for VQA tasks. We use a taxonomy of KGs to design a framework for identifying KGs for VQA questions. Using a question generation technique, we reduce the skew in the distribution of questions per KG category for the GQA dataset. In our future work, we aim to advance our question generation model and provide answers for generated questions. Additionally, we aim to generate more human-like question and to work towards resolving KGs. Our current work is an initial step towards understanding the cognitive skills need to advance AI models. \looseness=-1

\smallskip
\noindent
{\bf Acknowledgements:} We thank Dr. Wei-Lun Chao for suggestions on this work. The authors also acknowledge a seed grant from AFRL and OSU's Office of Research. \looseness=-1

{\small
\bibliographystyle{ieee_fullname}
\bibliography{egbib}
}

\end{document}